# PANDA-PLUS-Bench: A Clinical Benchmark for Evaluating Robustness of AI Foundation Models in Prostate Cancer Diagnosis

Joshua L. Ebbert[1], Dennis Della Corte[1,*]
[1]Department of Physics and Astronomy, Brigham Young University, 84602 Provo, USA
*Corresponding author: Dennis.DellaCorte@byu.edu

**Abstract**
Artificial intelligence foundation models are increasingly deployed for prostate cancer Gleason grading, where GP3/GP4 distinction directly impacts treatment decisions (active surveillance vs. intervention). However, these models may achieve high validation accuracy by learning specimen-specific artifacts rather than generalizable biological features, limiting real-world clinical utility. We introduce PANDA-PLUS-Bench, a curated benchmark dataset derived from expert-annotated prostate biopsies designed specifically to quantify this failure mode. The benchmark comprises nine carefully selected whole slide images from nine unique patients containing diverse Gleason patterns, with non-overlapping tissue patches extracted at both 512×512 and 224×224 pixel resolutions across eight augmentation conditions. Using this benchmark, we evaluate seven foundation models (Virchow, Virchow2, UNI, UNI2, Phikon, Phikon-v2, and HistoEncoder) on their ability to separate biological signal from slide-level confounders.

Our results reveal substantial variation in robustness across models: Virchow2 achieved the lowest slide-level encoding among large-scale models (slide ID accuracy: 81.0%) yet exhibited the second-lowest cross-slide accuracy (47.2%). HistoEncoder, trained specifically on prostate tissue, demonstrated the highest cross-slide accuracy (59.7%) and the strongest slide-level encoding (slide ID accuracy: 90.3%), suggesting tissue-specific training may enhance both biological feature capture and slide-specific signatures. All models exhibited measurable within-slide vs. cross-slide accuracy gaps, though the magnitude varied from 19.9 percentage points (HistoEncoder) to 26.9 percentage points (Phikon). We provide an open-source Google Colab notebook enabling researchers to evaluate additional foundation models against our benchmark using standardized metrics. PANDA-PLUS-Bench addresses a critical gap in foundation model evaluation by providing a purpose-built resource for robustness assessment in the clinically important context of Gleason grading.

## Introduction

Foundation models for computational pathology have emerged as powerful tools for automated diagnosis, trained on millions of whole slide images (WSIs) using self-supervised learning [1, 2]. These models generate feature embeddings adaptable to diverse downstream tasks, promising standardized, scalable diagnostic support. However, recent evidence suggests these models may encode spurious correlations (including scanner artifacts, staining protocols, and institutional processing differences) more strongly than the biological features they are intended to capture [3, 4].

The phenomenon of WSI-specific feature collapse, first characterized by Yun et al. [4], describes embedding spaces organized primarily by slide of origin rather than pathological features. De Jong et al. extended this observation, demonstrating that current foundation models encode medical center identity more strongly than cancer type across ten evaluated models [3]. Their proposed "Robustness Index," the ratio of biological to confounding feature encoding, exceeded 1.0 for only a single model (Virchow2 [5] at 1.2).

Beyond WSI-specific feature collapse, a hidden problem plagues computational pathology benchmarks: patient-level data leakage. When the same patient contributes multiple slides, common in biopsy procedures where 10–12 cores are standard, naive slide-level splitting allows patient identity to leak between train and test sets. Foundation models and downstream classifiers can exploit patient-specific signatures (tissue processing, immune infiltration patterns, stromal characteristics) rather than learning generalizable pathological features. This results in common failure modes when training classifiers on embeddings, shown in Figure 1.

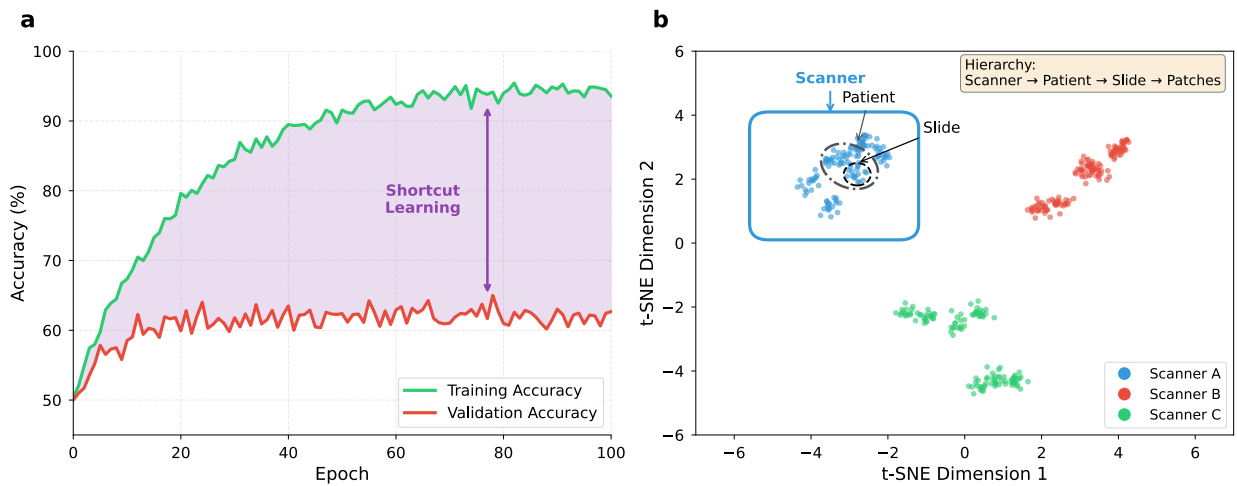

**Figure 1. Typical problems when training AI models on confounded embeddings.** a) Shortcut learning in foundation models and b) encoding of implicit hierarchical information in WSI embeddings.

This problem may contribute to inflated performance in competitions like PANDA, where winning solutions achieved QWK scores exceeding 0.93 [6]. Subsequent deployment studies have consistently shown degraded real-world performance, suggesting models learned dataset-specific shortcuts rather than robust Gleason pattern recognition. Critically, even "slide-level" cross-validation is insufficient when patient identity is encoded in tissue characteristics; the same patient's slides may cluster together in embedding space regardless of their Gleason content.

Despite growing recognition of these limitations, no standardized benchmark exists for systematically evaluating WSI-specific feature collapse and slide-level confounding across foundation models. Existing evaluations rely on heterogeneous datasets, inconsistent splitting strategies, and non-reproducible protocols. This gap impedes: (1) fair comparison across models, (2) tracking of progress in robustness, and (3) informed model selection for clinical applications.

We address this gap by introducing PANDA-PLUS-Bench, a purpose-built benchmark for evaluating foundation model robustness in Gleason pattern classification. Our contributions include (1) a curated multi-class benchmark dataset of nine WSIs from nine unique patients with expert pixel-level Gleason annotations, selected specifically for within-slide pattern diversity; (2) standardized evaluation protocols with defined metrics for quantifying slide-level confounding; (3) multi-resolution patches (512×512 and 224×224) enabling evaluation across model input requirements; (4) eight systematic augmentation conditions for assessing robustness to color and geometric variation; (5) comparative evaluation of seven foundation models (Virchow, Virchow2, UNI, UNI2, Phikon, Phikon-v2, HistoEncoder) spanning commercial and research licenses, general-purpose and tissue-specific training; and (6) an open-source evaluation toolkit (Google Colab notebook) enabling community benchmarking of additional models.

**Related Work**
*Pathology Foundation Models Under Evaluation*
We evaluate seven foundation models that span diverse training data, parameter size, and benchmarking performance. Training strategies, data scales, and licensing terms are described in Table 1. Prior Virchow benchmarking shows competitive performance in pan-cancer detection and other tasks. More recent benchmarks identify Virchow2 as among the top-performing models overall, particularly in representation of biological features over center-specific learning [7, 8]. The UNI pair of models, released under a research-only license, show strong performance on PANDA and offer diverse WSI training coverage [1]. Phikon, one of the earlier publicly released pathology foundation models, demonstrated that in-domain pre-training outperforms ImageNet pre-training [8]. Its successor, Phikon-v2, demonstrated performance on par with models trained on proprietary data, and the model won first place in the UCB-OCEAN Kaggle competition.

**Table 1.** Foundation models evaluated in this study. *Robustness Index from de Jong et al. [3] where available.

| Model | Architecture | Parameters | Training Data | License | Robustness Index* |
|---|---|---|---|---|---|
| Virchow [2] | ViT-H | 632M | 1.5M WSIs | Apache 2.0 | 0.93 |
| Virchow2 [5] | ViT-H | 632M | 3.1M WSIs | CC-BY-NC-ND 4.0 | 1.20 |
| UNI [1] | ViT-L | 307M | 100K WSIs | Research only | 0.88 |
| UNI2 [1] | ViT-L/H | ~632M | 350K+ WSIs | Research only | 0.93 |
| Phikon [9] | ViT-B | 86M | 40M tiles (TCGA) | Apache 2.0 | 0.84 |
| Phikon-v2 [10] | ViT-L | 300M | 460M tiles | Apache 2.0 | 0.74 |
| HistoEncoder [11] | ViT-S/B | ~85M | 48M tiles (prostate) | Apache 2.0 | --- |

The seven selected models support several orthogonal comparisons that contextualize their behavior in downstream evaluations. First, they contrast commercially deployable models (Virchow, Phikon, Phikon-v2, HistoEncoder) against research-restricted counterparts (UNI,

UNI2, Virchow2), and general-purpose encoders against the tissue-specific HistoEncoder. Second, they span distinct pre-training strategies (iBOT for Phikon versus DINOv2 for all others), a wide range of data scales from tens of millions of tiles to millions of WSIs, and both public (Phikon, Phikon-v2) and proprietary training corpora (Virchow, Virchow2, UNI), enabling analysis of how these factors influence robustness and transfer performance.

*Foundation Model Robustness: The Medical Center Problem*
Recent systematic evaluations have revealed fundamental robustness limitations in pathology foundation models. De Jong et al. evaluated ten publicly available pathology FMs and found that all models encode medical center information to a strong degree [3]. They introduced the "Robustness Index," a metric reflecting to what degree biological features dominate confounding features and found that only one model (Virchow2) achieved an index above 1.0, meaning biological features marginally dominated confounders. For all but one model, medical center origin could be predicted with 93-99% accuracy, and classification errors were not random, but attributable to same-center confounders.

A recent comprehensive analysis by Kömen et al. evaluated 20 foundation models and found robustness scores ranging from 0.463–0.877, with no model achieving full robustness [12]. They found a correlation ($\rho = 0.692$) between training data size and robustness and demonstrated that even combining data robustification (stain normalization) with representation robustification (batch correction) could not eliminate performance drops across medical centers.

*Data Leakage in Digital Pathology: A Systematic Problem*
The problem of data leakage in digital pathology has been systematically documented. Bussola et al. demonstrated that predictive performance estimates can be inflated by up to 41% when tiles from the same patient appear in both training and validation sets [13]. Their work established that patient-wise splitting is essential. This issue is then exacerbated when fractions or subregions of slides (referred to as tiles or patches) are considered.

This concern extends to AI competitions. While the PANDA challenge organizers implemented blinded external validation on held-out cohorts from the US and EU, achieving QWK scores of 0.862 and 0.868 respectively [6], questions remain about whether patient-level separation was enforced during the competition's development phase, when participants had access to training data. Competition environments create incentives for participants to exploit any available shortcuts, including patient-level correlations [14].

More broadly, the gap between retrospective AI performance and real-world clinical efficacy remains pronounced. Recent reviews emphasize that, although many models achieve favorable results in retrospective studies, rigorous prospective, multi-center evaluation is still rare, which in turn constrains their impact on clinical guidelines and everyday decision-making [15]. Shortcomings of available datasets may contribute to persistent performance gaps [16, 17]. Our benchmark addresses the slide-level confounding problem by explicitly measuring within-slide vs. cross-slide generalization gaps, a necessary (though not sufficient) condition for robust clinical deployment.

*Gleason Grading as a Robustness Test Case*
Prostate cancer Gleason grading represents an ideal test case for robustness evaluation because the GP3/GP4 distinction directly influences treatment decisions (active surveillance vs. intervention). Inter-observer variability ($\kappa = 0.43$–$0.67$) establishes realistic performance bounds [18, 19], and architectural features distinguishing patterns are subtle and easily confounded by technical artifacts [20].

**Methods**
*Source Data: PANDA-PLUS*
PANDA-PLUS comprises 546 whole slide images from the PANDA challenge [6], re-annotated at pixel level through a hierarchical annotation pipeline over approximately three years. Annotations were performed by trained undergraduate pre-medical students, supervised by senior reviewers, with final verification by a board-certified pathologist (>30 years of experience). Annotators were recruited from the Brigham Young University pre-medical undergraduate population through the AI in Medicine Association and trained in the fundamentals of prostate histology, the Gleason grading system, gland morphology, and digital annotation techniques [21, 22]. Each glandular structure received a Gleason pattern classification (Benign, GP3, GP4, or GP5) at 20× magnification using QuPath [23].

*Benchmark Slide Selection Criteria*
To enable rigorous within-slide vs. cross-slide comparison, we applied stringent selection criteria. More than 30 patches of majority benign, GP3, and GP4 tissue were required per slide with complete, pixel-level annotations verified by an expert pathologist. Slides with significant artifacts, focus issues, or tissue damage were excluded. Given that each slide in the PANDA dataset represents an independent biopsy specimen, our cross-slide evaluation inherently tests generalization beyond individual tissue samples. The within-slide vs. cross-slide accuracy gap measured therefore represents the model's ability to generalize learned Gleason patterns to new specimens rather than memorizing specimen-specific features. From 546 PANDA-PLUS images, nine slides met all criteria (Table 2).

**Table 2.** Benchmark slide characteristics.

| Slide ID | Benign | GP3 | GP4 | GP5 | Total | Institution |
|---|---|---|---|---|---|---|
| 27 | 71 | 135 | 136 | — | 342 | Radboud |
| 101 | 113 | 37 | 226 | — | 376 | Radboud |
| 391 | 50 | 154 | 37 | — | 241 | Radboud |
| 416 | 32 | 48 | 75 | 58 | 213 | Radboud |
| 2571 | 55 | 93 | 85 | — | 233 | Radboud |
| 3010 | 80 | 136 | 57 | — | 273 | Radboud |
| 3333 | 68 | 44 | 223 | — | 335 | Radboud |
| 3501 | 50 | 47 | 280 | — | 377 | Radboud |
| 3524 | 50 | 68 | 262 | — | 380 | Radboud |
| **Total** | 569 | 762 | 1,381 | 58 | 2,770 | — |

*Patch Processing*

Non-overlapping patches were extracted from annotated tissue regions at two resolutions: 224×224 pixels at 20× magnification (0.4862 µm/pixel), being standard input size for most foundation models, and 512×512 pixels at 20× magnification (0.4862 µm/pixel), providing higher context for models supporting larger inputs. The 224x224 patches are non-overlapping subsets of the 512x512 patches. Each patch received a ground-truth label based on the dominant (>50%) Gleason pattern from pixel-level annotations, on average comprising 83.0% (224x224) and 89.3% (512x512) of the patch. Patches with no dominant class were excluded. Each patch was processed under eight augmentation conditions to assess robustness to technical variation, shown in Table 3.

**Table 3.** Augmentation conditions in PANDA-PLUS-Bench.

| Condition | Description | Rationale |
|---|---|---|
| Baseline | ImageNet normalization only | Control condition |
| ColorJitter | Brightness±0.2, contrast±0.2, saturation±0.3, hue±0.04 | Simulates lighting variation |
| Grayscale | Complete color removal | Isolates texture/structure features |
| Gaussian Noise | Additive noise ($\sigma=0.05$) | Simulates image quality variation |
| Heavy Geometric | Rotation±180°, H/V flips | Tests rotation invariance |
| Macenko | Stain normalization to reference [24] | Standard stain normalization |
| HED Augmentation | H/E channel perturbation (±20%) | Simulates staining variation |
| Combined Aggressive | All augmentations applied sequentially | Maximum perturbation |

*Dataset Statistics and Availability*

The complete benchmark dataset, including an executable Google Colab notebook, is available at https://github.com/dellacortelab/panda-plus-bench and https://huggingface.co/datasets/dellacorte/PANDA-PLUS-Bench.

*Embedding Extraction*

For each of the seven foundation models, embeddings were extracted from all image patches under every augmentation setting according to model-specific preprocessing protocols. Virchow and Virchow2 produced 2560-dimensional representations obtained by concatenating the CLS token with the mean of patch tokens, whereas UNI and UNI2 yielded 1024-dimensional CLS-token embeddings, all using ImageNet normalization. Phikon generated 768-dimensional CLS-token embeddings from its ViT-B backbone, and Phikon-v2 generated 1024-dimensional CLS-token embeddings from its ViT-L backbone, both with ImageNet normalization, while HistoEncoder features were derived following the official repository guidelines. All embeddings were cached to permit efficient computation of evaluation metrics across experimental conditions.

*Primary Metrics: Within-Slide vs. Cross-Slide Accuracy*

The primary evaluation metrics contrasted within-slide and cross-slide performance. Within-slide accuracy was computed by treating each slide independently, applying a stratified 80/20 train-test split using k-nearest neighbors (k=5) classification, and averaging accuracy across slides. Cross-slide accuracy was estimated using leave-one-slide-out cross-validation, training on eight slides and testing on the held-out slide, then averaging accuracy over all held-out slides. The accuracy gap was defined as within-slide minus cross-slide accuracy, with larger gaps interpreted as evidence of stronger slide-level confounding. Additional metrics are described in Table 4.

**Table 4.** Robustness metrics in PANDA-PLUS-Bench.

| Metric | Definition | Interpretation |
|---|---|---|
| **kNN Same-Slide Fraction** | Fraction of k=50 nearest neighbors from same slide | Expected by chance: 11.1% |
| **Slide ID Accuracy** | 5-fold CV accuracy predicting slide from embedding | High accuracy (>>11.1% chance) indicates strong slide encoding |
| **Silhouette Score** | Mean silhouette coefficient computed using either slide IDs or class labels as cluster assignments | Near-zero class values indicate overlapping class structure; positive slide values indicate residual slide-level clustering. |

The benchmark includes standardized visualizations: t-SNE projections colored by class and by slide, centroid distance matrices showing within-slide vs. cross-slide distances, per-slide accuracy heatmaps identifying challenging cases, robustness radar plots for multi-model comparison, and accuracy hierarchy diagrams illustrating the leakage problem.

**Results**

*Multi-Model Comparison: Primary Metrics*

**Table 5.** Primary classification metrics across foundation models (Baseline condition).

| Model | Within-Slide Acc | Cross-Slide Acc | Gap |
|---|---|---|---|
| Virchow | 0.715 | 0.508 | 0.208 |
| Virchow2 | 0.718 | 0.472 | 0.202 |
| UNI | 0.767 | 0.519 | 0.249 |
| UNI2 | 0.715 | 0.476 | 0.240 |
| Phikon | 0.746 | 0.477 | 0.269 |
| Phikon-v2 | 0.722 | 0.472 | 0.250 |
| HistoEncoder | 0.796 | 0.597 | 0.199 |

Classification performance varied substantially across the seven foundation models (Table 5). Within-slide accuracy ranged from 0.715 (Virchow, UNI2) to 0.796 (HistoEncoder), while cross-slide accuracy ranged from 0.472 (Virchow2, Phikon-v2) to 0.597 (HistoEncoder) under baseline conditions. HistoEncoder achieved both the highest cross-slide accuracy (0.597) and the smallest accuracy gap (0.199), while Phikon exhibited the largest gap (0.269) despite mid-range within-slide performance (0.746).

All seven models demonstrated positive accuracy gaps, indicating consistently higher performance when training and testing on patches from the same slide compared to leave-one-slide-out cross-validation. The Virchow and Virchow2 models, which share identical architectures but differ in training data scale (1.5M vs. 3.1M WSIs), showed similar within-slide accuracy (0.715 vs. 0.718) but divergent cross-slide performance (0.508 vs. 0.472).

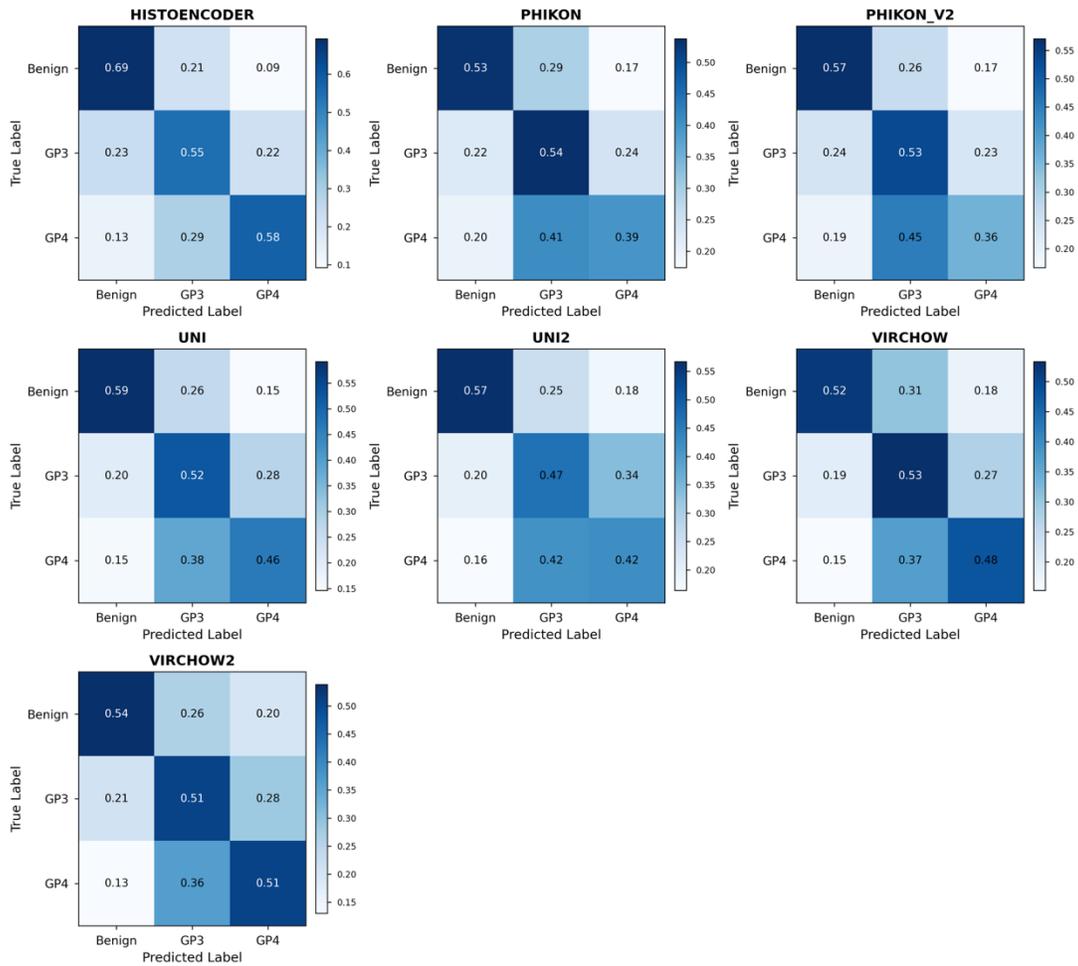

**Figure 2.** Confusion matrices by model for predicted versus true Gleason pattern scores. Darker blue indicates greater alignment.

Confusion matrices (Figure 2) revealed distinct error patterns across models. Benign-versus-tumor discrimination was strongest for all models, while GP3-GP4 boundary errors were more frequent. HistoEncoder showed proportionally fewer cross-class errors between benign and tumor tissue compared to general-purpose models.

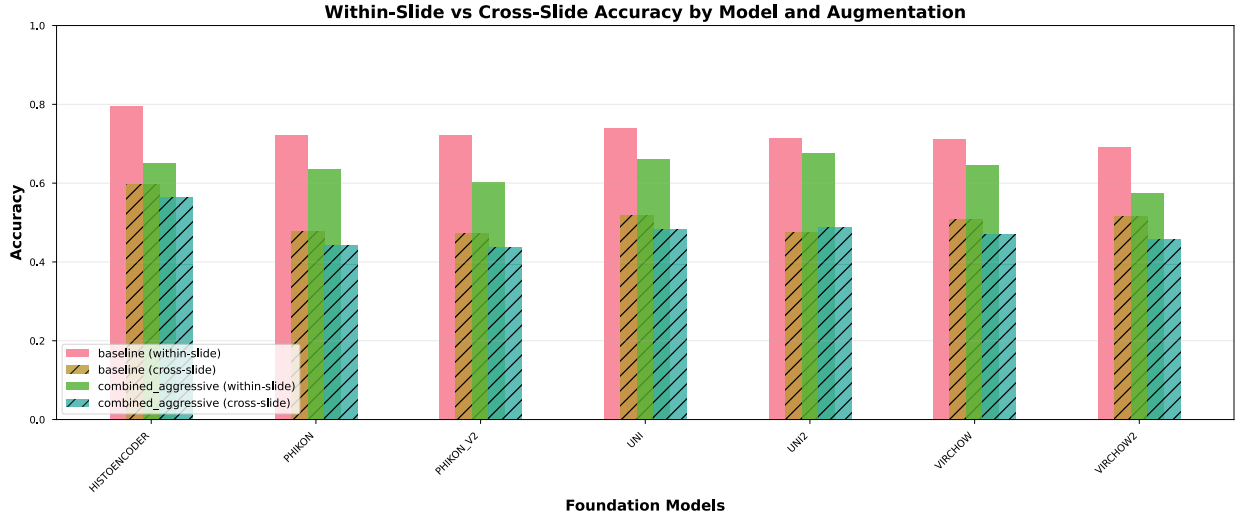

**Figure 3.** Within-slide versus cross-slide accuracy by model for baseline and combined-aggressive augmentation conditions.

Across all seven models, baseline within-slide accuracy exceeded cross-slide accuracy and combined-aggressive augmentation reduced both metrics (Figure 3). HistoEncoder maintained the highest cross-slide performance under both conditions, while the Virchow models showed the largest absolute drops in cross-slide accuracy when moving from baseline to combined-aggressive augmentation. UNI2 was the sole model to exhibit a higher cross-slide accuracy under the combined-aggressive augmentation relative to baseline.

*Multi-Model Comparison: Robustness Metrics*
**Table 6.** Robustness metrics across foundation models (Baseline condition).

| Model | Silh. Score (Class) | Silh. Score (Slide) | Slide ID Acc | kNN Same-Slide |
|---|---|---|---|---|
| Virchow | -0.016 | 0.018 | 0.807 | 0.579 |
| Virchow2 | 0.001 | 0.021 | 0.810 | 0.553 |
| UNI | 0.002 | 0.028 | 0.846 | 0.621 |
| UNI2 | 0.000 | 0.018 | 0.850 | 0.614 |
| Phikon | -0.002 | 0.041 | 0.846 | 0.638 |
| Phikon-v2 | -0.002 | 0.042 | 0.850 | 0.622 |
| HistoEncoder | 0.015 | 0.091 | 0.903 | 0.619 |

Slide-level encoding strength varied markedly across models (Table 6). Slide ID prediction accuracy ranged from 0.807 (Virchow) to 0.903 (HistoEncoder), substantially exceeding the chance baseline of 0.111 (1/9 slides). The kNN same-slide fraction metric, which measures the proportion of k=50 nearest neighbors originating from the same slide, ranged from 0.553 (Virchow2) to 0.638 (Phikon), compared to an expected chance value of 0.111.

Silhouette scores computed using class labels were near-zero or slightly negative for all models (range: -0.016 to 0.015), indicating substantial overlap in the embedding space between Gleason pattern classes. In contrast, silhouette scores computed using slide IDs were consistently positive (range: 0.018 to 0.091), with HistoEncoder showing the highest slide-based clustering (0.091).

Virchow2 achieved the lowest slide ID accuracy (0.810) among large models and the lowest kNN same-slide fraction (0.553) overall, while simultaneously showing near-zero class silhouette scores (0.001) and low slide silhouette scores (0.021). HistoEncoder exhibited the strongest slide-level encoding (slide ID accuracy 0.903, silhouette score 0.091) alongside the highest cross-slide classification performance.

Figure 4 radar plots compare models across four metrics simultaneously: cross-slide accuracy, within-slide accuracy, kNN same-slide fraction (inverted), and slide ID accuracy (inverted. Models with larger polygons demonstrate more balanced robustness profiles.

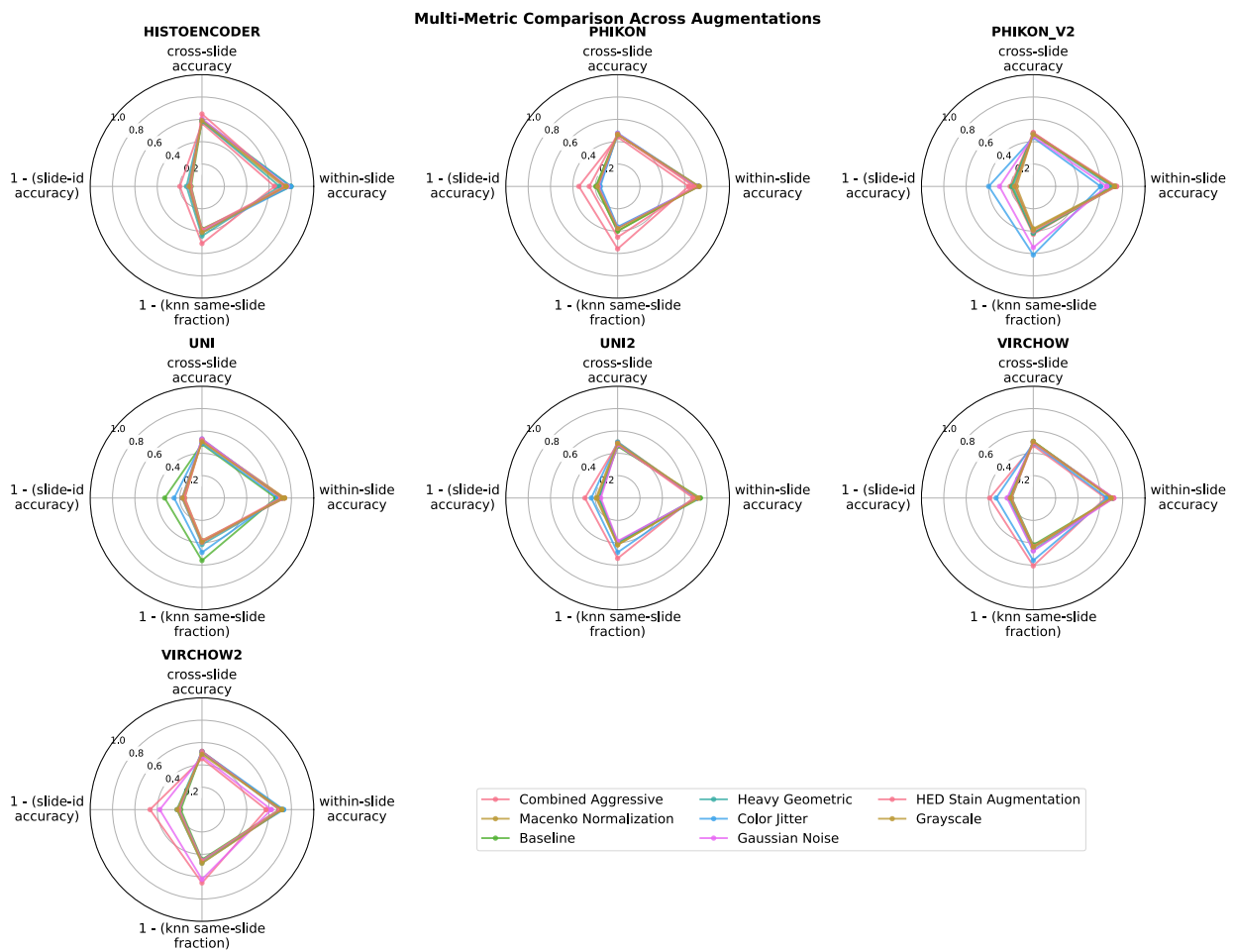

**Figure 4.** Multi-metric robustness profiles for seven pathology foundation models across eight augmentation conditions. Radar plots show within-slide accuracy, cross-slide accuracy, 1 – slide ID accuracy, and 1 – kNN same-slide fraction for each model, with larger polygons indicating higher cross-slide performance and weaker slide-specific encoding under the corresponding augmentation.

*Effect of Augmentation on Model Rankings*
Model performance and rankings shifted under different augmentation conditions (Figure 4). Under baseline conditions, HistoEncoder ranked first in cross-slide accuracy (0.597), followed by UNI (0.519) and Virchow (0.508). Grayscale conversion, which eliminates all color information, reduced cross-slide accuracy for all models, with reductions ranging 0.5-4 percentage points.

Heavy geometric augmentation (180° rotation, horizontal and vertical flips) produced variable effects across models. ColorJitter augmentation, simulating lighting variation through brightness, contrast, saturation, and hue perturbations, decreased cross-slide accuracy by 0.5-3 percentage points relative to baseline across models.

Stain normalization methods (Macenko, HED augmentation) showed model-specific responses. Phikon and Phikon-v2, both trained on public TCGA data with standardized staining protocols, exhibited different sensitivity patterns to stain perturbations compared to models trained on proprietary multi-institutional data. The combined aggressive augmentation condition, applying all transformations sequentially, resulted in the largest performance degradation across all models, with cross-slide accuracy declining to 0.437-0.565.

Accuracy gap magnitude (within-slide minus cross-slide) remained relatively stable across augmentation conditions for most models, varying by less than 0.06 for all but HistoEncoder and Phikon-v2. HistoEncoder maintained the smallest gap (0.09-0.21) across all eight augmentation conditions.

*Resolution Effects (224×224 vs. 512×512)*
Among the evaluated models, HistoEncoder was the only architecture operating on 512×512 patches, while all other foundation models were assessed at 224×224 resolution. Under these conditions, HistoEncoder achieved the highest within-slide (0.796) and cross-slide (0.597) accuracy, the smallest accuracy gap (0.199), and the highest slide ID prediction accuracy (0.903) among all models in the benchmark.

*Training Method Comparison*
Phikon (iBOT pretraining) and Phikon-v2 (DINOv2 pretraining) provide a controlled comparison of self-supervised learning strategies, both trained on public TCGA data. Phikon achieved higher cross-slide accuracy (0.477) compared to Phikon-v2 (0.472) under baseline conditions, despite Phikon-v2's larger model size (300M vs. 86M parameters) and substantially larger training dataset (460M tiles vs. 40M tiles).

Phikon-v2 demonstrated lower within-slide accuracy (0.722 vs. 0.746) and a smaller accuracy gap (0.250 vs. 0.269) relative to Phikon. Slide-level encoding metrics were nearly identical between the two models: slide ID accuracy of 0.846 (Phikon) versus 0.850 (Phikon-v2), and silhouette scores for slide clustering of 0.041 versus 0.042.

Under grayscale augmentation, Phikon-v2 showed a near identical cross-slide performance compared to a slight performance decrease by Phikon (-1.9 percentage points). Under all stain normalization conditions except combined aggressive and heavy geometric (Macenko, HED, gaussian noise, etc.), both models showed similar robustness, with cross-slide accuracy shifting less than 2.7 percentage points.

*Embedding Space Visualization*
HistoEncoder embeddings showed the most pronounced slide-based clustering. Virchow2 embeddings exhibited the least visible slide-based structure in t-SNE space, corresponding to its lowest slide ID accuracy (0.810) among models with comparable parameter counts. Class separation, measured qualitatively by visual cluster distinctness in t-SNE plots (Figure 5), was weakest for the GP3-GP4 boundary across all models. Benign tissue showed partial separability from tumor classes (GP3, GP4), but with considerable overlap.

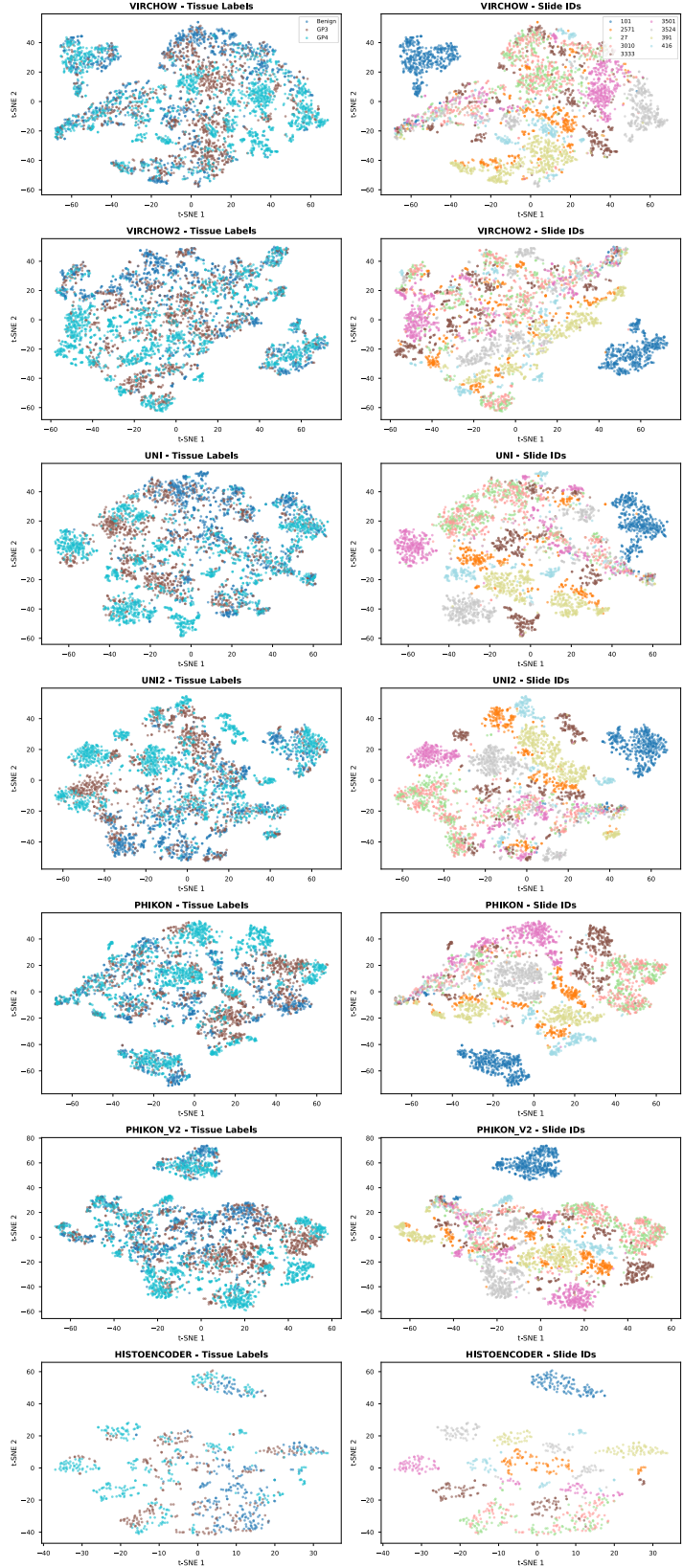

**Figure 5.** t-SNE visualizations of foundation model embedding spaces colored by tissue label (left column) and slide ID (right column) for Virchow, Virchow2, UNI, UNI2, Phikon, Phikon-v2, and HistoEncoder under the baseline condition. Each panel shows two-dimensional projections of patch embeddings, enabling qualitative assessment of biological class separation versus slide-specific clustering for each model.

*Per-Slide Analysis*

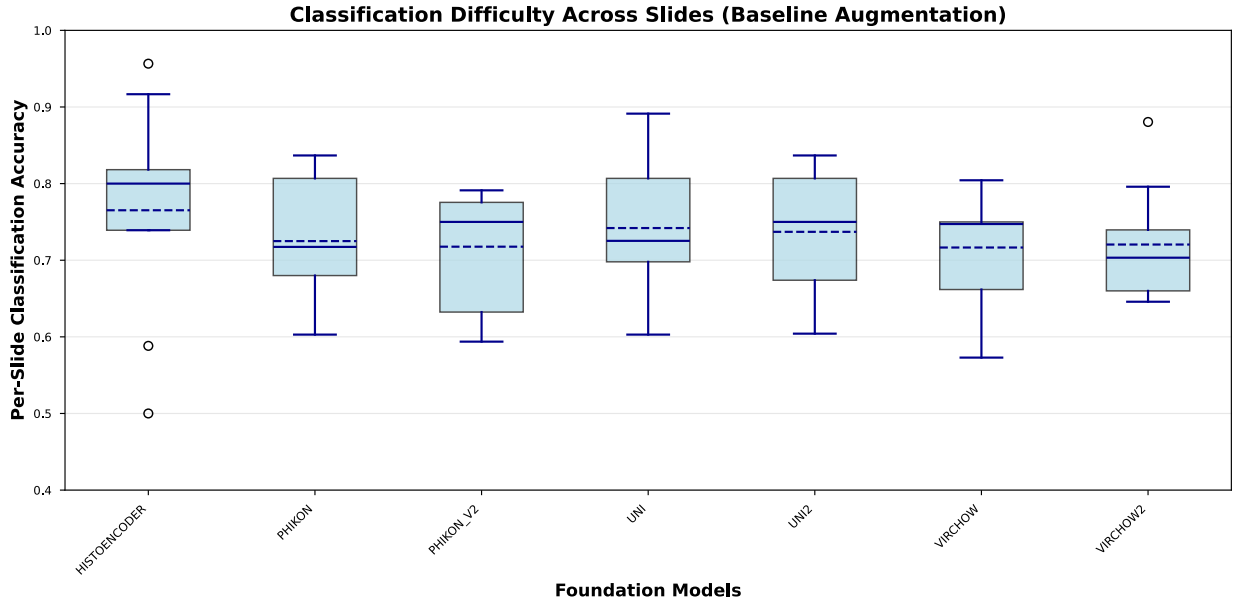

**Figure 6.** Per-slide classification difficulty across foundation models under baseline augmentation. Box plots show the distribution of leave-one-slide-out cross-slide accuracies for each model across the nine benchmark slides, with solid lines indicating the median and dashed lines indicating the mean per-slide accuracy.

Across slides, median cross-slide accuracy ranged from approximately 0.7 to 0.8 depending on the model, with substantial variability in performance between slides (Figure 6). HistoEncoder showed the highest median per-slide accuracy and the narrowest interquartile range with several low-accuracy outliers corresponding to particularly challenging slides, whereas general-purpose models exhibited wider spreads.

## Discussion
*Model Rankings and Trade-offs*
Across the seven evaluated models, HistoEncoder achieved the highest cross-slide accuracy under the baseline condition and exhibited the smallest within–cross accuracy gap, indicating comparatively stronger generalization across held-out slides in this prostate-specific task. General-purpose foundation models trained on large, heterogeneous datasets (Virchow, Virchow2, UNI, UNI2, Phikon, Phikon-v2) displayed lower cross-slide accuracy than HistoEncoder and uniformly positive gaps, with Phikon showing the largest gap despite mid-range within-slide performance. Licensing status did not align with robustness: both commercially deployable and research-only models appeared across the spectrum of cross-slide accuracy and gap magnitudes, with no consistent separation by license category in primary or robustness metrics. Training data scale likewise showed no monotonic relationship with robustness in this benchmark; for example, Virchow2, trained on more WSIs than Virchow, achieved similar within-slide accuracy but lower cross-slide accuracy in the baseline setting, and Phikon-v2, trained on many more tiles than Phikon, did not surpass it on cross-slide accuracy. On the prostate-specific task, HistoEncoder outperformed general-purpose encoders in both within- and cross-slide accuracy while also presenting the highest slide ID accuracy, reflecting a distinct trade-off profile relative to general-purpose counterparts. Relative to its predecessor, Phikon-v2's public-data-only training did not confer an advantage on the primary cross-slide metric under baseline conditions; the two models showed closely matched slide-encoding indicators and similar responses to stain-focused augmentations.

Our results provide complementary evidence to the cross-center robustness findings of de Jong et al.[3]. While their Robustness Index measures the dominance of biological features (cancer type) over institutional confounders across medical centers, our metrics quantify slide-level confounding within a single institution. Notably, Virchow2's highest Robustness Index (1.20) in their multi-center evaluation corresponds to the lowest slide ID accuracy (81.0%) among comparable models in our benchmark, suggesting consistency in reduced confounding across evaluation contexts. However, the relationship between their Robustness Index and our cross-slide accuracy is non-monotonic: Virchow (Index: 0.93) achieved higher cross-slide accuracy (50.8%) than Virchow2 (47.2%), despite Virchow2's superior Robustness Index. This dissociation suggests that reduced encoding of confounders is necessary but insufficient for robust task-specific performance, and that different types of confounding (institutional vs. slide-level) may require distinct mitigation strategies.

*The Persistence of Slide-Level Confounding*
All models demonstrated higher within-slide than cross-slide performance, and slide ID could be predicted from embeddings well above chance for every model, indicating persistent slide-level signatures in representation space. These findings suggest that, even when classification accuracy is moderate at the patch level, embeddings retain information specific to individual slides, as further supported by positive slide-based silhouette scores and t-SNE visualizations showing distinct slide clusters across models.

*Augmentation as Diagnostic, Not Remedy*
Our results demonstrate that post-hoc augmentation during embedding extraction does not remediate WSI-specific feature collapse. Even grayscale conversion, which eliminates all color information, preserves high slide ID classification accuracy, indicating that slide identity is encoded through texture, compression artifacts, and structural patterns beyond color.
This finding has important implications: augmentation applied during inference cannot undo shortcuts learned during pre-training. Models achieving better robustness on our benchmark likely benefit from training-time interventions (diverse data sources, stain augmentation during pre-training, architectural choices).

*Benchmark Limitations and Future Extensions*
This study focuses on a single organ site (prostate), includes primarily one contributing institution, and evaluates nine slides selected for stringent annotation quality and class coverage, which may not capture the full spectrum of specimen and acquisition variability. From a clinical deployment perspective, our benchmark evaluates specimen-level robustness but not patient-level robustness (multiple specimens per patient), institution-level robustness (cross-center generalization), or temporal robustness (performance stability as protocols evolve). While specimen-level generalization is necessary for clinical utility, it is not sufficient; models must also generalize across these additional hierarchies before deployment in diverse clinical settings.

*Implications for Competition Design and Benchmark Reporting*
The results underscore the necessity of rigorous data splitting strategies when evaluating pathology foundation models. While our benchmark measures slide-level generalization the consistent accuracy gaps observed (20-27 percentage points) demonstrate that even specimen-level confounding poses substantial challenges for current models.

Competition organizers and benchmark creators should implement hierarchical splitting strategies that match their data structure: patient-level splits when multiple samples per patient exist, slide-level splits when slides represent independent specimens, and institution-level splits when evaluating cross-center generalization. Leaderboards should report performance at each relevant level of hierarchy separately (e.g., within-patient vs. cross-patient, within-slide vs. cross-slide, within-institution vs. cross-institution), as divergence between these metrics reveals susceptibility to shortcut learning at different granularities.

Benchmarks should incorporate multi-institutional data with explicit stratification to guard against center-specific shortcuts and should invest in expert re-annotation where label quality is uncertain, with inter-observer agreement statistics made explicit. We propose a robustness-first reporting standard that requires: (1) demonstration that test performance generalizes beyond the finest grain of data collection (patches → slides → patients → institutions → scanners), (2) quantification of accuracy gaps at each hierarchical level, and (3) measurement of slide/patient/institution encoding strength using metrics such as those presented here (slide ID accuracy, kNN same-slide fraction, silhouette scores). Such standards would better align evaluation with intended deployment contexts and expose models that achieve high aggregate accuracy through memorization of specimen-specific or institutional signatures rather than robust biological feature learning.

*Clinical Implications*
Our findings have direct implications for deploying AI-assisted Gleason grading in clinical workflows. The 20-27 percentage point accuracy gaps between within-slide and cross-slide performance suggest that models validated on retrospective datasets may underperform on new patient specimens by similar margins. For HistoEncoder, the best-performing model, 59.7% cross-slide accuracy falls below the lower bound of inter-observer agreement ($\kappa = 0.43$-$0.67$), indicating current foundation models cannot yet match pathologist-level reliability when generalizing to new specimens.

The persistent slide-specific encoding across all models (81-90% slide ID accuracy) poses risks for clinical deployment. If deployed models rely on specimen-specific signatures, they may fail when scanning protocols, tissue processing, or staining methods change – common occurrences in clinical laboratories. This failure mode would be particularly problematic for the GP3/GP4 boundary, where our confusion matrices show the highest error rates and where clinical consequences (treatment selection) are most significant.

The tissue-specific HistoEncoder model's superior cross-slide performance (59.7% vs. 47-52% for general-purpose models) suggests that organ-specific foundation models may offer better clinical utility than general-purpose alternatives for diagnostic tasks with narrow pathological scope. However, its high slide encoding (90.3% slide ID accuracy) indicates that even specialized training cannot eliminate specimen-specific confounding without explicit architectural or training interventions.

These findings suggest that current foundation models require task-specific fine-tuning with robust validation protocols before clinical deployment for Gleason grading. Institutions should prioritize models demonstrating strong cross-specimen performance on internal validation cohorts over those achieving high accuracy on public benchmarks, where data leakage may inflate performance estimates.

*Recommendations for Practitioners*
Before deployment, practitioners should establish evaluation protocols that match their intended use case. At minimum, this requires splits that prevent data leakage at the finest grain of data collection—slide-level separation when slides represent independent specimens, patient-level separation when multiple samples per patient are collected, and institution-level separation for cross-center deployment. Our benchmark demonstrates that even slide-level separation, a less stringent requirement than patient-level separation, reveals substantial performance gaps (20-27 pp) across state-of-the-art foundation models.

Accuracy alone is insufficient for robustness evaluation. Practitioners should complement classification metrics with structural assessments of the embedding space: (1) accuracy gaps between within-group and cross-group performance quantify the magnitude of confounding, (2) slide/patient ID prediction accuracy reveals how strongly specimen identity is encoded, (3) kNN same-slide/patient fraction measures local clustering by non-biological variables, and (4) silhouette scores computed using both biological labels and confounding variables assess the relative strength of desired versus undesired structure. Models showing high slide ID accuracy (>70%) despite moderate classification performance should be treated as potentially encoding shortcuts rather than biological features.

Reported competition or retrospective performance should be interpreted cautiously unless validation methodology is transparent. Specifically, practitioners should verify: (1) whether data splitting prevented leakage at all hierarchical levels (patch/tile, slide, patient, institution), (2) whether reported metrics reflect cross-group generalization or within-group performance, and (3) whether augmentation strategies were applied during training (potentially beneficial) versus only during inference (minimally effective, as our results demonstrate). The gap between PANDA competition performance (QWK ~0.93) and subsequent real-world deployment studies suggests that shortcuts exploitable during development may not transfer to clinical settings.

When baseline robustness is inadequate for the intended application, practitioners should consider: (1) tissue-specific foundation models when available (HistoEncoder achieved 59.7% cross-slide accuracy vs. 47-52% for general-purpose models in our prostate-specific task), (2) task-specific fine-tuning with carefully designed splits and augmentation strategies applied during training, and (3) ensemble approaches that explicitly combine models with different robustness profiles. Evaluation protocols should mirror the deployment setting, including institutional diversity, scanner heterogeneity, and staining protocol variation when relevant. Existing models should be audited for potential hierarchical leakage before clinical use. This audit should: (1) verify that training data splits prevented all forms of leakage (patient, institution, scanner), (2) measure cross-group generalization performance on held-out data matching the deployment context, (3) quantify slide/patient/institution encoding strength using structural metrics, and (4) assess robustness to expected technical variation through systematic augmentation testing. Models failing these audits require retraining with corrected splits or architectural modifications before deployment.

**Conclusion**

We introduce PANDA-PLUS-Bench, a purpose-built benchmark for evaluating WSI-specific feature collapse in pathology foundation models. Using this benchmark, we evaluated seven foundation models representing diverse training strategies, data scales, and licensing terms, and found that all models exhibited substantial slide-level encoding (slide ID prediction accuracy 81-90% vs. 11% chance), with accuracy gaps between within-slide and cross-slide performance ranging from 20-27 percentage points. Tissue-specific training (HistoEncoder) achieved the highest cross-slide accuracy (59.7%) but also the strongest slide-specific signatures, while general-purpose models trained on larger datasets showed more variable robustness profiles with no consistent relationship between training scale and cross-slide generalization.

Our results demonstrate that current pathology foundation models, despite strong within-slide performance, exhibit persistent susceptibility to slide-level confounding that cannot be remediated through inference-time augmentation alone. Post-hoc color normalization, geometric transformations, and even complete grayscale conversion preserved high slide ID prediction accuracy, indicating that slide identity is encoded through texture, compression artifacts, and structural patterns learned during pre-training. The benchmark and evaluation toolkit are available at https://github.com/dellacortelab/panda-plus-bench and https://huggingface.co/datasets/dellacorte/PANDA-PLUS-Bench, enabling standardized robustness evaluation as foundation models continue to evolve. For clinical adoption, our results suggest that current foundation models require: (1) explicit validation on held-out specimens from the target institution, (2) monitoring for performance degradation when scanning or processing protocols change, and (3) consideration of tissue-specific models when available. The benchmark and evaluation toolkit we provide enable institutions to perform these validations systematically before deployment. We encourage the community to benchmark additional models using our open-source framework and contribute to understanding and ultimately solving the challenge of slide-level confounding in computational pathology.